\PassOptionsToPackage{table,xcdraw}{xcolor}
\documentclass{article} 
\usepackage{iclr2025_conference,times}

\usepackage{amsmath,amsfonts,bm}

\def\eqref#1{equation~\ref{#1}}

\def\1{\bm{1}}

\DeclareMathAlphabet{\mathsfit}{\encodingdefault}{\sfdefault}{m}{sl}
\SetMathAlphabet{\mathsfit}{bold}{\encodingdefault}{\sfdefault}{bx}{n}

\usepackage{hyperref}
\usepackage{url}

\usepackage{tcolorbox}
\tcbuselibrary{listingsutf8}

\usepackage{graphicx}

\definecolor{myalpha}{HTML}{D62728} 
\definecolor{myentropy}{HTML}{1F77B4}
\usepackage{booktabs}

\definecolor{mauve}{RGB}{152, 138, 172}
\definecolor{lightmauve}{RGB}{186, 175, 199}

\definecolor{lavender}{RGB}{200, 190, 230}
\definecolor{amethyst}{RGB}{180, 160, 210}
\definecolor{sakura}{RGB}{230, 200, 220}

\definecolor{softblue}{RGB}{160, 180, 210}
\definecolor{softgreen}{RGB}{160, 210, 180}

\definecolor{lightblue}{RGB}{190, 210, 230} 

\definecolor{ForestGreen}{HTML}{228B22}

\usepackage{enumitem}
\setlist[itemize]{noitemsep, topsep=0pt}

\usepackage{wrapfig}
\usepackage{subcaption}
\usepackage{caption}

\usepackage{tikz}
\usetikzlibrary{arrows.meta, positioning}
\usepackage{amsmath}

\usepackage{mathtools} 
\usepackage{textgreek}

\makeatletter
\def\thanks#1{\protected@xdef\@thanks{\@thanks
        \protect\footnotetext{#1}}}
\makeatother
\title{Reasoning with Exploration:\\An Entropy Perspective}

\author{Daixuan Cheng\textsuperscript{\textalpha,\textbeta,\textepsilon}\textsuperscript{\dag}\thanks{Email: \texttt{daixuancheng6@gmail.com}
 \textsuperscript{\dag}Core Contributors.\textsuperscript{\ddag} Corresponding Authors.}~~Shaohan Huang\textsuperscript{\textbeta}\textsuperscript{\dag}~~Xuekai Zhu\textsuperscript{\textgamma}\textsuperscript{\dag}~~Bo Dai\textsuperscript{\textepsilon}\\[3pt]
\textbf{Wayne Xin Zhao\textsuperscript{\textalpha}\textsuperscript{\ddag}~~Zhenliang Zhang\textsuperscript{\textepsilon}\textsuperscript{\ddag}~~Furu Wei\textsuperscript{\textbeta}}
\\[6pt]
\textsuperscript{\textalpha}RUC\quad\textsuperscript{\textbeta}MSRA\quad\textsuperscript{\textgamma}SJTU\quad\textsuperscript{\textepsilon}BIGAI
\vspace{-15pt}}

\iclrfinalcopy 
\begin{document}

\maketitle

\begin{abstract}
Balancing exploration and exploitation is a central goal in reinforcement learning (RL). Despite recent advances in enhancing large language model (LLM) reasoning, most methods lean toward exploitation, and increasingly encounter performance plateaus. In this work, we revisit entropy—a signal of exploration in RL—and examine its relationship to exploratory reasoning in LLMs. Through empirical analysis, we uncover positive correlations between high-entropy regions and three types of exploratory reasoning actions: (1) pivotal tokens that determine or connect logical steps, (2) reflective actions such as self-verification and correction, and (3) rare behaviors under-explored by the base LLMs. Motivated by this, we introduce a minimal modification to standard RL with only one line of code: augmenting the advantage function with an entropy-based term. Unlike traditional maximum-entropy methods which encourage exploration by promoting uncertainty, we encourage exploration by promoting longer and deeper reasoning chains. Notably, our method achieves significant gains on the Pass@$K$ metric—an upper-bound estimator of LLM reasoning capabilities—even when evaluated with extremely large $K$ values, pushing the boundaries of LLM reasoning.
\end{abstract}

\begin{figure}[!htb]
    \centering
    \vspace{-10pt}
\includegraphics[width=0.85\linewidth]{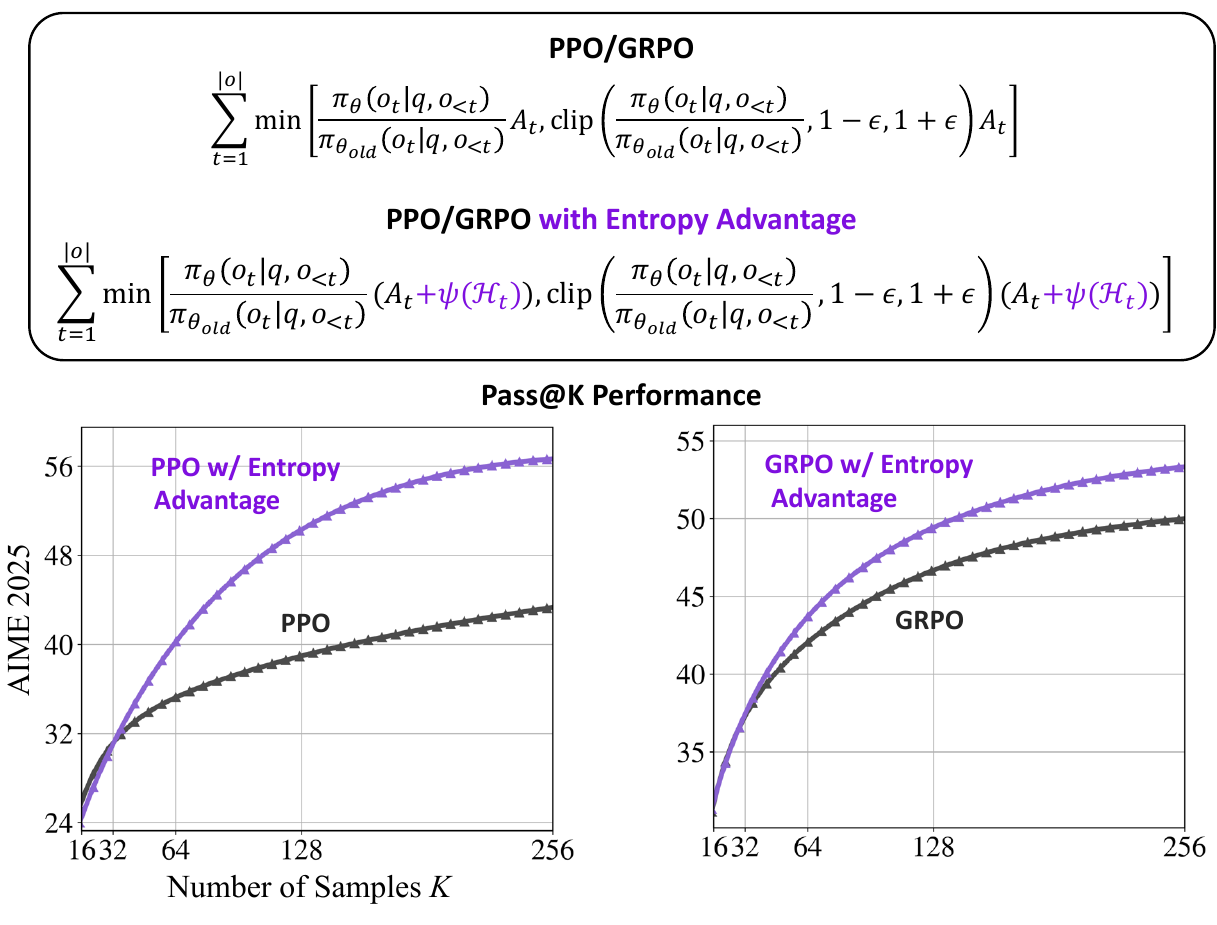}
    \vspace{-7pt}
    \caption{\textbf{Top}: We augment the advantage in PPO~\citep{PPO} or GRPO~\citep{grpo} with a minimal per-token entropy-based term. 
\textbf{Bottom}: Our entropy-based advantage effectively encourages exploratory reasoning in LLMs, achieving superior Pass@$K$ performance even with extremely large $K$ values.
}\label{fig:intro}
\end{figure}

\section{Introduction}
\vspace{-3pt}
Recent reinforcement learning methods for large language models (LLMs), particularly those using verifiable rewards (RLVR;~\citealp{rlvr}), typically rely on signals that reflect output accuracy to guide training. These approaches have proven effective in enhancing reasoning by reinforcing correct outputs and discouraging incorrect ones~\citep{deepseekr1}. However, as training progresses under purely accuracy-driven objectives, these benefits often diminish. LLMs tend to converge on narrow and over-optimized behaviors, gradually losing their incentive to explore alternative strategies. This lack of exploration weakens the model’s capacity for sustained, multi-step reasoning, causing performance to plateau or even regress, especially in complex or underspecified settings~\citep{dapo,entropymechanism}.

In traditional RL, exploration plays a vital role alongside exploitation by encouraging the policy model to explore alternative strategies and avoid overfitting. A common metric for measuring exploration is entropy, which quantifies uncertainty in the policy’s action distribution~\citep{SAC,InverseRL}. Motivated by this, we investigate the relationship between entropy and exploratory reasoning in LLMs, and uncover strong correlations: (1) Pivotal tokens that guide or connect reasoning steps—such as \texttt{first}, \texttt{because}, and \texttt{however}—consistently exhibit higher entropy; (2) Reflective actions~\citep{rethinkingreflection}, such as self-verification and error correction, tend to emerge under high-entropy conditions; (3) During RL training, rare or under-explored solutions also coincide with elevated entropy. Together, these findings suggest entropy can be a valuable signal for recognizing exploratory reasoning behaviors in LLMs.

Based on these findings, we propose incorporating entropy as an auxiliary term to encourage exploratory reasoning during RL training. While traditional maximum entropy methods encourage exploration by promoting uncertainty~\citep{pgq}, our approach takes a different path to balance exploration and exploitation: we introduce a clipped, gradient-detached entropy term into the advantage function of standard RL algorithms. Clipping ensures that the entropy term neither dominates nor reverses the sign of the original advantage, while gradient detachment preserves the original optimization direction. This design amplifies exploratory reasoning behaviors that emerge under uncertainty while maintaining the original policy gradient flow. Moreover, because of the intrinsic tension between entropy and confidence, the entropy-based term naturally diminishes as confidence increases—encouraging exploration in early stages while avoiding over-encouragement as training progresses. Furthermore, our method is extremely simple, requiring only one line of code to seamlessly integrate into existing RLVR training pipelines~\citep{verl}.

We validate our method on mainstream RLVR algorithms, GRPO~\citep{grpo} and PPO~\citep{PPO}, and observe distinct benefits. First, it amplifies exploratory reasoning behaviors—such as the use of pivotal tokens and reflective actions—by decreasing the policy's uncertainty at these decision points. Second, it encourages the generation of longer, more exploratory responses without increasing the repetition rate, enabling coherent multi-step reasoning. Consequently, our method consistently improves Pass@$1$ accuracy across different benchmarks. We further increase the number of attempts $K$ per question to evaluate Pass@$K$—a metric recently regarded as an upper-bound estimator of reasoning capability~\citep{RLLimit}. As shown in Figure~\ref{fig:intro}, our method achieves substantial improvements even at large $K$, pushing the boundaries of LLM reasoning. In summary, the key contributions of this work are as follows:
\begin{itemize}[leftmargin=*]
\itemsep0em
\item We investigate and reveal a strong correlation between entropy and exploratory reasoning in LLMs, showing that pivotal tokens, reflective actions, and rare behaviors emerge with higher entropy.
\item We propose a minimal yet effective method that augments the standard RL advantage with a clipped, gradient-detached entropy term, encouraging exploration by fostering longer and deeper reasoning chains while preserving the original policy optimization direction.
\item We validate our method on mainstream RLVR algorithms: GRPO and PPO, achieving consistent gains on Pass@$1$ and substantial improvements on Pass@$K$, pushing the boundaries of reasoning.
\end{itemize}

\begin{figure}[!htb]
    \centering
\includegraphics[width=\linewidth]{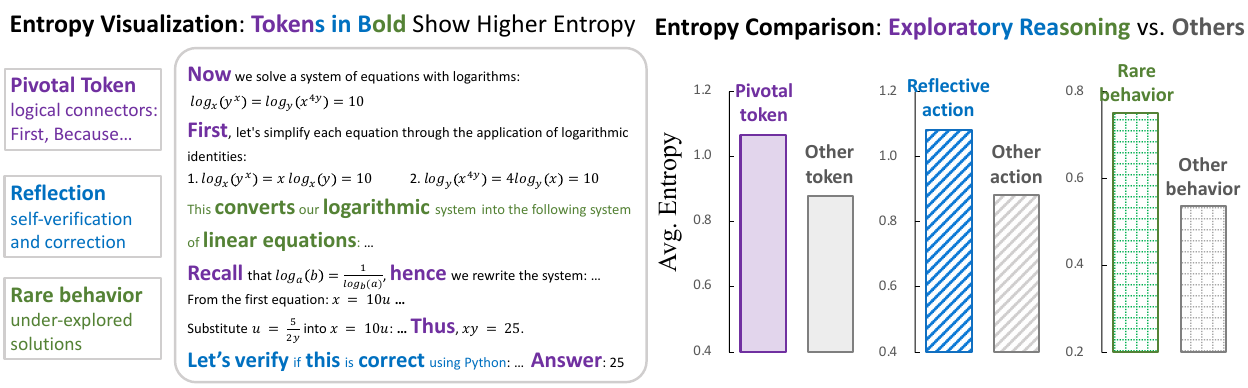}
\vspace{-15pt}
    \caption{\textbf{Entropy Visualization and Comparison between Exploratory Reasoning and Others}. We categorize tokens/actions/behaviors based on their role in the reasoning process. In the visualization, tokens with higher entropy appear in bold and larger sizes, with colors denoting different reasoning roles. In the comparison, we show average entropy values across different categories.}
    \label{fig:analysis}
\end{figure}

\section{Preliminary Analysis: Entropy and Exploratory Reasoning}\label{sec:Preliminary Analysis}
We examine \textit{entropy}—a core signal of exploration in RL~\citep{equivalence,SAC,bridging}—and its relationship with exploratory reasoning in LLMs. We begin by visualizing token-level entropy in the responses of Qwen2.5-Base-7B~\citep{qwen2.5} on mathematical reasoning tasks~\citep{AIME}. As shown in Figure~\ref{fig:analysis}, we observe that high-entropy tokens consistently correspond to different reasoning dynamics compared to low-entropy ones. Based on this observation, we categorize exploratory reasoning-related content—including both tokens and sentences—to support the following analysis\footnote{We also conduct a parallel analysis using a fine-tuned LLM—DeepSeek-R1-Distill-Qwen-1.5B~\citep{deepseekr1}—on code reasoning tasks, yielding consistent observations (see Appendix~\ref{app:Extended Preliminary Analysis}).}.

\paragraph{Pivotal Tokens}
Figure~\ref{fig:analysis} shows that pivotal reasoning tokens (e.g., \texttt{first}, \texttt{recall}, \texttt{thus}) tend to have higher entropy. These tokens serve as logical connectors, marking decision points where the model determines the flow and structure of reasoning. To quantify this observation, we compute the average entropy of commonly occurring pivotal tokens across all responses and compare it to that of the remaining tokens. These include causal terms (e.g., \texttt{because}, \texttt{therefore}), contrastive markers (e.g., \texttt{however}, \texttt{although}), sequential terms (e.g., \texttt{first}, \texttt{then}), and reasoning verbs (e.g., \texttt{suggest}, \texttt{demonstrate}). Results on the right of Figure~\ref{fig:analysis} confirm a statistically significant increase in entropy for these pivotal tokens. Similar observations have also been noted in concurrent work~\citep{beyond82rule,demystifying}, where such tokens are referred to as forking tokens or information peaks.

\paragraph{Reflective Actions}
Reflection is a form of meta-cognition that involves examining generated information, evaluating the underlying reasoning, and adapting future behavior accordingly~\citep{rethinkingreflection}. In this work, we focus on self-reflection, where the model assesses and comments on its own outputs. This is illustrated in the visualization in Figure~\ref{fig:analysis}, where the LLM assigns higher entropy to sentences such as ``Let's verify if this is correct...''. 

\begin{wrapfigure}[15]{r}{0.43\textwidth}
\vspace{-25pt}
\centering
\includegraphics[width=0.43\textwidth]{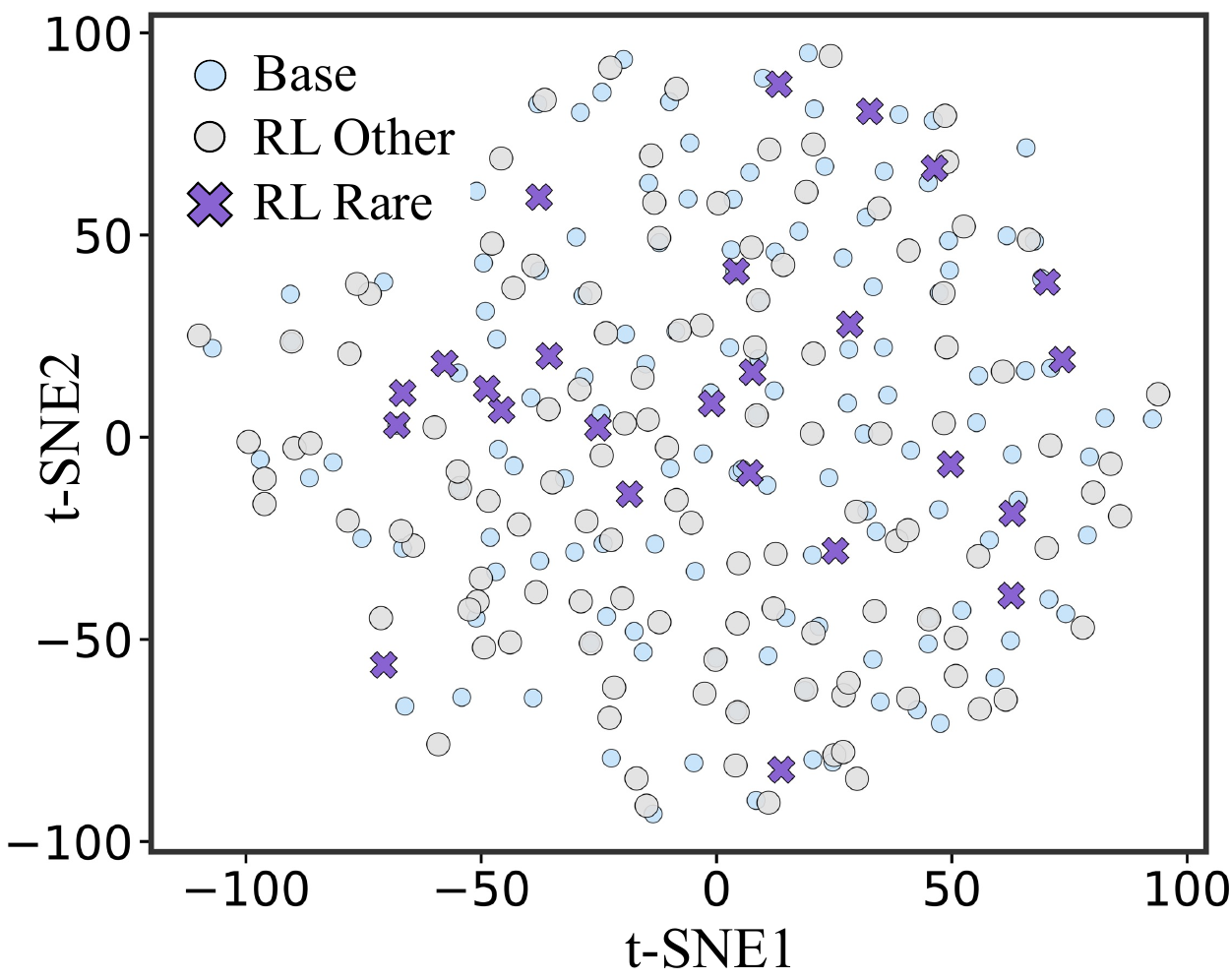}
\vspace{-20pt}
\caption{\textbf{Behavior Clustering}. t-SNE projection of response embeddings. Base denotes the pre-RL model outputs; RL Other and RL Rare represent common and rare behaviors after RL, respectively.} 
\label{fig:clustering}
\end{wrapfigure}
To quantify this behavior, we segment each response into sentences, compute the average entropy for each one, and use regular expressions to identify reflective actions—specifically, sentences containing keywords such as “verify” or “check”. As shown in the comparison in Figure~\ref{fig:analysis}, these reflective sentences consistently exhibit higher average entropy, suggesting that self-reflection tends to occur under greater uncertainty. To the best of our knowledge, this is the first analysis linking entropy to self-reflection in LLMs.

\paragraph{Rare Behaviors Emergent During RL}
We further examine whether under-explored or emergent behaviors—those rarely exhibited by the base model—are associated with distinct entropy patterns during RL. In the visualization (Figure~\ref{fig:analysis}), such behaviors include converting logarithmic systems into systems of linear equations, which are less frequently observed in the base model's outputs. To quantify this, we perform RL training on the base model (see Section~\ref{sec:exp setting} for configurations), and define rare behaviors as sentences that semantically isolated from the base model’s output distribution. We embed all response sentences using SBERT~\citep{sbert} and, for each RL-generated sentence, compute the average distance to its $k=5$ nearest base-model neighbors. Sentences in the top 10\% of this distance metric are labeled as rare. Behavior clusters are visualized in Figure~\ref{fig:clustering}. As shown in the comparison in Figure~\ref{fig:analysis}, these rare behaviors exhibit higher entropy, revealing a strong correlation between semantic novelty and predictive uncertainty.

\section{Method}
Our analysis reveals a strong correlation between entropy and exploratory reasoning in LLMs, motivating us to actively encourage high-entropy actions during training. To this end, we propose an advantage shaping method that augments the per-token advantage with a term based on its entropy. This entropy-based term serves as a robust, self-regulating signal that guides learning without altering the original gradient flow of the base RL algorithm.

Let \( q \) denote a question sampled from a dataset \( \mathcal{D} \), and let \( o = (o_1, o_2, \dots, o_{|o|}) \) be the corresponding output response generated by a policy model \( \pi_\theta \). Our method is compatible with mainstream policy optimization algorithms such as Proximal Policy Optimization (PPO;~\citealp{PPO}) and Group Relative Policy Optimization (GRPO;~\citealp{grpo}). We begin by briefly reviewing these methods before introducing our entropy-based advantage shaping method.

\subsection{RL Baselines: PPO and GRPO}

\paragraph{Proximal Policy Optimization (PPO)}
PPO optimizes the policy by maximizing the following clipped surrogate objective:
\begin{align}
    \mathcal{J}_{\mathrm{PPO}}(\theta) = \mathbb{E}_{q \sim \mathcal{D},\, o \sim \pi_{\theta_{\text{old}}}(O\mid q)} \left\{
    \sum_{t=1}^{|o|}
    \min \left[ 
        \rho_t(\theta) A_t,\,
        \mathrm{clip}(\rho_t(\theta), 1 - \epsilon_{\mathrm{low}}, 1 + \epsilon_{\mathrm{high}}) A_t 
    \right]
    \right\},
\end{align}
where \( \rho_t(\theta) = \frac{\pi_\theta(o_t\mid q, o_{<t})}{\pi_{\theta_{\text{old}}}(o_t \mid q, o_{<t})} \) denotes the likelihood ratio between the current and old policy models, and \( A_t \) is the advantage, typically computed using Generalized Advantage Estimation (GAE;~\citealp{gae}. We omit the length normalization term, as our implementation adopts a token-level policy loss without per-response normalization. The loss is averaged across all tokens in a training batch to mitigate implicit length bias~\citep{drgrpo,simplerl}. The clipping range \( \epsilon_{\mathrm{low}} \) and \( \epsilon_{\mathrm{high}} \) stabilizes policy updates by preventing excessively large changes. While standard PPO uses symmetric clipping (i.e., \( \epsilon_{\mathrm{low}} = \epsilon_{\mathrm{high}} \)), recent work~\citep{dapo} suggests that slightly increasing \( \epsilon_{\mathrm{high}} \) can help avoid entropy collapse.

The gradient of the PPO objective is (we omit $\min$ and $\mathrm{clip}$ operations under the single-update-per-rollout assumption~\citep{grpo}):

\begin{align}\label{equ:ppo}
\nabla_\theta \mathcal{J}_{\mathrm{PPO}}(\theta) 
&= \mathbb{E}_{q \sim \mathcal{D},\, o \sim \pi_{\theta_{\text{old}}}(O\mid q)} \left[
\sum_{t=1}^{|o|}  A_t 
 \nabla_\theta \log \pi_\theta(o_t\mid q, o_{<t}) \right].
\end{align}

\paragraph{Group Relative Policy Optimization (GRPO)}
GRPO is an alternative to GAE-based PPO that avoids learning a separate value function by using the average reward of multiple sampled outputs, produced in response to the same question, as the baseline. Formally, for each question $ q $, a group of $ G $ outputs \( \{o_1, o_2, \dots, o_G\} \)  is sampled from the old policy $ \pi_{\theta_{\text{old}}} $, a reward model is then used to score the outputs, yielding $G$ rewards $\{r_1, r_2, \dots, r_G\}$ correspondingly. These scores are then normalized as:
\begin{align}
\tilde{r}_i = \frac{r_i - \text{mean}(\{r_1, r_2, \dots, r_G\} )}{\text{std}(\{r_1, r_2, \dots, r_G\} )}.
\end{align}

Recently, GRPO has been widely used in outcome-supervised settings~\citep{deepseekr1}, where the normalized reward is assigned at the end of each output \( o_i \), and every token in \( o_i \) receives the same advantage, i.e., \( {A}_{i,t} = \tilde{r}_i \) . The policy is then optimized using the PPO objective in Equation~\ref{equ:ppo} with these group-relative advantages. A KL penalty term between the trained policy and a reference policy may be added to the loss~\citep{kl_approx}.

\subsection{Encouraging Exploratory Reasoning via Entropy-based Advantage}

\paragraph{Entropy-Based Advantage Shaping} To encourage exploratory reasoning, we propose an entropy-guided advantage shaping method. The key idea is to inject an entropy-based term into the advantage function during policy optimization.

For each token \( o_t \) in an output \( o \), the entropy of the current policy over the vocabulary \( \mathcal{V} \) is:
\begin{align}\label{equ:entropy}
\mathcal{H}_t = -\sum_{v \in \mathcal{V}} \pi_\theta(v \mid q, o_{<t}) \log \pi_\theta(v \mid q, o_{<t}).
\end{align}
We then define an entropy-based advantage term \(\psi(\mathcal{H}_t)\) and use it to shape the advantage:
\begin{align}\label{equ:bonus}
\psi(\mathcal{H}_t) = \min\left(\textcolor{myalpha}{\alpha} \cdot \textcolor{myentropy}{\mathcal{H}^{\mathrm{detach}}_t},~ \frac{|A_t|}{\textcolor{ForestGreen}{\kappa}} \right), \quad \text{where  }\textcolor{myalpha}{\alpha} > 0 \text{  and  }\textcolor{ForestGreen}{\kappa} > 1,
\end{align}
\begin{align}  
A^{\mathrm{shaped}}_t = A_t + \psi(\mathcal{H}_t).
\end{align}
Here, \textcolor{myalpha}{\( \alpha \)} is the scaling coefficient, and \textcolor{ForestGreen}{\( \kappa \)} controls the clipping threshold. This clipping ensures that the entropy-based term \( \psi(\mathcal{H}_t) \leq \frac{|A_t|}{\textcolor{ForestGreen}{\kappa}} \), so it does not dominate the advantage. Moreover, when \( A_t < 0 \), this constraint ensures that adding the entropy-based term does not reverse the sign of the advantage—thus preserving the original optimization direction. Crucially, the entropy term \textcolor{myentropy}{$\mathcal{H}^{\mathrm{detach}}_t$} is detached from the computational graph during backpropagation, acting as a fixed offset to the original advantage. This adjusts the magnitude of the update without altering the gradient flow. As a result, the policy gradient retains a format similar to that of PPO in Equation~\ref{equ:ppo}, where only the advantage $A_t$ is replaced with the shaped one:
\begin{align}
\nabla_\theta \mathcal{J}^{\mathrm{shaped}}_{\mathrm{PPO}}(\theta) 
&= \mathbb{E}_{q \sim \mathcal{D},\, o \sim \pi_{\theta_{\text{old}}}(O\mid q)} \left[
\sum_{t=1}^{|o|} (A_{t} + \psi(\mathcal{H}_{t}))
 \nabla_\theta \log \pi_\theta(o_t\mid q, o_{<t}) \right].
 \label{equ:our_gradient}
\end{align}

Our shaping method can be seamlessly integrated into existing RL training pipelines using only a single line of code. Specifically, after computing the advantages with PPO or GRPO, we add the entropy-based advantage term before calculating the policy loss, as follows\footnote{To be specific, this corresponds to a one-line code insertion in the \texttt{update\textunderscore policy} function of \texttt{verl/workers/actor/dp\textunderscore actor.py} file when using the veRL framework~\citep{verl}.}:

\begin{tcolorbox}[
    title=Entropy-Based Advantage Shaping (PyTorch Implementation),
    colback=white,
    colframe=lavender,
    coltitle=gray!40!black,
    fonttitle=\bfseries,
    arc=1mm,
    boxrule=0.6mm,
    left=1mm,   
    right=1mm,   
    top=1mm,     
    bottom=1mm,  
]
\ttfamily
\# Compute advantages as in PPO or GRPO \\
adv = compute\_advantages(...) \\[6pt]
\# Apply entropy-based term for advantage shaping \\
adv += torch.min(\textcolor{myalpha}{alpha} * \textcolor{myentropy}{entropy.detach()}, 
adv.abs()/\textcolor{ForestGreen}{kappa}) \\[6pt]
\# Use the shaped advantages to compute the policy loss \\
loss = compute\_policy\_loss(adv, ...)
\end{tcolorbox}

\begin{wrapfigure}[19]{r}{0.43\textwidth}
\vspace{-3pt}
\centering
\includegraphics[width=0.43\textwidth]{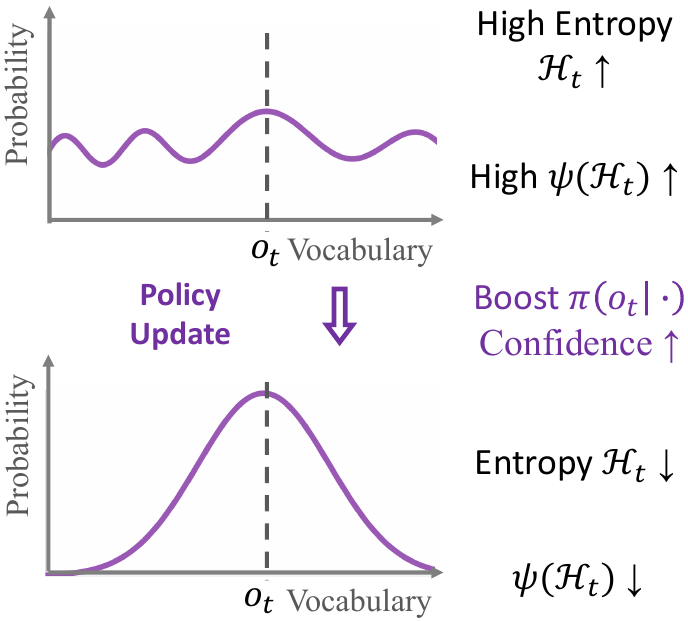}
\vspace{-15pt}
\caption{\textbf{Dynamics of Entropy-Based Advantage}. High entropy initially largely amplifies the advantage, accelerating confidence gain and leading to reduced entropy-based shaping in subsequent steps.}
\label{fig:bonus_loop}
\end{wrapfigure}

\paragraph{Robustness of Entropy-Based Advantage: Avoiding Over-Encouragement} Prior work~\citep{still3} attempts to enhance reasoning by rewarding the policy based on the frequency of reasoning-like tokens, but this leads to reward hacking—the policy model repeatedly generates such tokens to exploit the reward without performing true reasoning. In contrast, our method naturally avoids such over-encouragement due to the intrinsic tension between entropy and confidence. As shown in Figure~\ref{fig:bonus_loop}, our method initially assigns high advantage to tokens with high-entropy distributions but gradually reduces the entropy-based advantage as model confidence increases over training iterations.

Formally, let \(k\) denote the training iteration and \(t\) denote the token position within the output response. The policy model parameters are updated via gradient ascent:
\begin{align}
    \theta_{k+1} = \theta_k + \eta \, \nabla_\theta \mathcal{J}(\theta_k),
\end{align}
where \(\eta\) is the learning rate, and the policy gradient \(\nabla_\theta \mathcal{J}(\theta_k)\) (Equation~\ref{equ:our_gradient}) uses the shaped advantage \(A^{\mathrm{shaped}}_{k,t} = A_{k,t} + \psi(\mathcal{H}_{k,t})\) which is positively correlated with the detached entropy \(\textcolor{myentropy}{\mathcal{H}^{\mathrm{detach}}_{k,t}}\) (Equation~\ref{equ:bonus}). When the original advantage \(A_{k,t} > 0\), higher entropy leads to a stronger update on the selected token \(o_t\), largely increasing its likelihood \(\pi_\theta(o_t \mid \cdot)\) and thus sharpening the output distribution. According to the entropy definition in Equation~\ref{equ:entropy}, a sharper distribution lowers entropy, which in turn reduces the entropy-based advantage  \(\psi(\mathcal{H}_{t}) \) and weakens subsequent updates. This self-regulating effect is empirically validated in Figure~\ref{fig:bonus_ratio}.

\paragraph{Comparison with Entropy Regularization}
In traditional RL, it is common to add an entropy regularizer to the gradient to prevent the policy from becoming overly deterministic~\citep{pgq}. Practically, this means adding an entropy loss term to the policy loss. To clarify the distinction between our method and entropy regularization, we present a comparison in Table~\ref{tab:entropy_comparison}.

Entropy regularization explicitly adds an entropy term \( \sum_t \mathcal{H}_{t} \) to the objective, scaled by a coefficient \( \beta \). Since \( \mathcal{H}_{t} \) depends on the current policy \( \pi_\theta \), this introduces an additional gradient component \( \nabla_\theta \mathcal{H}_{t} \), encouraging higher-entropy policies during training.

In contrast, our method modifies the advantage function by adding a clipped entropy term \textcolor{myentropy}{\( \mathcal{H}_{t}^{\mathrm{detach}} \)}, which is detached from the computation graph. As a result, \( \nabla_\theta \textcolor{myentropy}{\mathcal{H}_t^{\mathrm{detach}}} = 0 \), and the entropy term influences optimization only through the adjusted advantage values. Thus, our method preserves the original RL optimization dynamics. This makes it fundamentally distinct from---and even orthogonal to---entropy regularization.

\begin{table}[ht]
\centering
\resizebox{\textwidth}{!}{\begin{tabular}{lll}
\toprule
 & \textbf{Entropy Regularization} & \textbf{Entropy-Based Adv. Shaping} \\
\midrule
Training Objective & 
\( \mathcal{J} = \mathcal{J}_{\mathrm{PPO}} + \beta \sum_t \mathcal{H}_t \) & 
\( \mathcal{J} = \mathcal{J}_{\mathrm{PPO}}(A^{\mathrm{shaped}}_t) \) \\[6pt]

Policy Gradient & 
\( \sum_t A_t \nabla_\theta \log \pi_\theta(o_t) + \beta \sum_t \nabla_\theta \mathcal{H}_t \) & 
\( \sum_t A^{\mathrm{shaped}}_t \nabla_\theta \log \pi_\theta(o_t) \) \\[6pt]

Entropy Gradient Flow & 
\( \nabla_\theta \mathcal{H}_t \neq 0 \) & 
\( \nabla_\theta \textcolor{myentropy}{\mathcal{H}_t^{\mathrm{detach}}} = 0 \) \\
\bottomrule
\end{tabular}}
\caption{Comparison of gradient behavior between entropy regularization and our entropy-based advantage shaping. We present simplified expressions that omit PPO's $\min$/$\mathrm{clip}$ operations and batch normalization. \( \mathcal{J}_{\mathrm{PPO}}(A^{\mathrm{shaped}}_t) \) denotes the PPO objective computed with shaped advantages.}
\label{tab:entropy_comparison}
\end{table}

\section{Experiment Settings}\label{sec:exp setting}
\paragraph{Backbone Models}
We conduct experiments on two base models: the general-purpose Qwen2.5-Base-7B~\citep{qwen2.5} and its domain-adapted variant Qwen2.5-Math-Base-7B~\citep{qwen2.5math}. We also initially attempted RL training from Llama-series LLMs~\citep{llama3} using vanilla GRPO or PPO, but observed that the LLMs abandoned intermediate reasoning chains within just a few training iterations. This observation aligns with \citet{Cognitivebehaviors} that Llama LLMs inherently lack reasoning behaviors and likely require pre-training on reasoning traces prior to RL training. 

\paragraph{RL Training Configuration} Our training data are sourced from DAPO~\citep{dapo}. We use output reward that assigns +1 for correct final answers and -1 otherwise. We conduct experiments on GRPO and PPO using the veRL framework~\citep{verl}. To build strong baselines, we adopt several techniques from DAPO and VAPO~\citep{VAPO}, including Clip-Higher, Token-level Loss, Critic-Pretraining, and Group-Sampling. Detailed hyperparameters are in Appendix~\ref{app:Experiment Settings}. Building on these RL baselines, we apply our proposed entropy-based advantage. We fix \textcolor{ForestGreen}{$ \kappa $} to $2$ throughout all experiments, and set \textcolor{myalpha}{$ \alpha $} to $0.4$ for GRPO and $0.1$ for PPO.

\paragraph{Evaluation Benchmarks and Metrics}
We evaluate on AIME 2025/2024~\citep{AIME}, AMC 2023~\citep{AMC} and MATH500~\citep{Math500}, using a rollout temperature of $0.6$, a maximum response length of $8K$ tokens, and top-$p$ sampling with $p=0.95$. Each dataset is evaluated multiple times, and we report the average Pass@$1$ accuracy. Following~\citet{RLLimit}, we also assess reasoning ability boundaries using the Pass@$K$ metric: for each question, if at least one of $K$ sampled model outputs passes verification, Pass@$K=1$; otherwise $0$. To mitigate variance, we adopt the unbiased estimation method proposed by~\citet{unbiasedpassk}. For the small and challenging benchmark {AIME 2024/2025} (30 examples per year), we scale \( K \) to a large value of 256. For the larger and less challenging benchmarks {AMC 2023} (83 examples) and {MATH500} (500 examples), we set \( K = 128 \) and \( K = 16 \), respectively, because LLMs already achieve near-perfect results with small \( K \), and their size makes large \( K \) computationally expensive.

\section{Results}

\begin{figure}[!htb]
    \centering
\includegraphics[width=\linewidth]{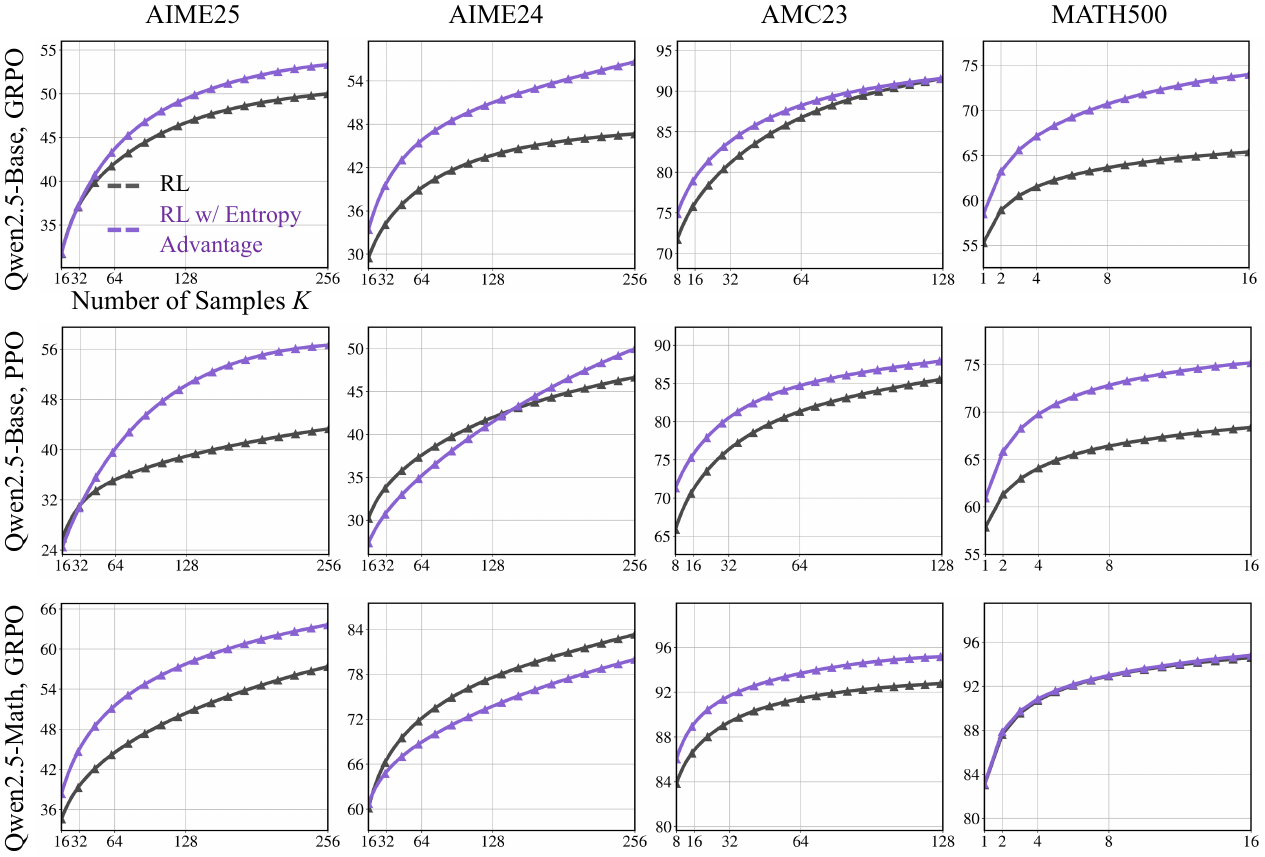}
    \vspace{-15pt}
    \caption{\textbf{Pass@K Performance} of the LLMs with different RL algorithms.}
    \label{fig:passk}
\end{figure}

As shown in Table~\ref{tab:main_results}, our method consistently outperforms the baselines across benchmarks and RL algorithms, achieving superior average performance even compared to strong existing approaches~\citep{prime,drgrpo,gpg}. Moreover, this advantage extends to Pass@$K$—a metric for estimating the reasoning capacity of LLMs. As shown in Figure~\ref{fig:passk}, our method continues to deliver improvements even at large $K$ values, where most baselines plateau.

On benchmarks such as AIME2024, AMC2023, and MATH500, we observe a similar phenomenon reported in~\citet{RLLimit}: although RL-trained models consistently outperform their base models in terms of average Pass@$1$ performance, the base models can surpass RL-finetuned ones in Pass@$K$ when $K$ becomes sufficiently large. This indicates that conventional RL fine-tuning may inadvertently limit the exploratory capacity of the model. Our method effectively mitigates this issue. Notably, on AIME2025—the most challenging benchmark in our evaluation, released after the training data cutoff of the base models—{our method not only outperforms the RL baselines but also exceeds the performance ceiling of the base model}. This highlights the potential of our approach to break through the inherent limitations of base models and push the boundaries of LLM reasoning.

\begin{table}[!htb]
\resizebox{\textwidth}{!}{%
\begin{tabular}{l@{\hspace{5pt}}cccccccc}
\toprule
                         & \multicolumn{2}{c}{AIME25} & \multicolumn{2}{c}{AIME24} & \multicolumn{2}{c}{AMC23} & \multicolumn{2}{c}{MATH500} \\ \cmidrule(lr){2-3} \cmidrule(lr){4-5} \cmidrule(lr){6-7} \cmidrule(lr){8-9}
                         & \small\textit{Pass@256}      & \small\textit{Pass@1}     & \small\textit{Pass@256}      & \small\textit{Pass@1}     & \small\textit{Pass@128}     & \small\textit{Pass@1}     & \small\textit{Pass@16}       & \small\textit{Pass@1}      \\ \cmidrule(lr){1-9}
\hspace{-4pt}\textit{{Qwen2.5-Base}}      & 50.0        & 2.2      & 66.7        & 5.2      & 90.4       & 28.3     & 88.8        & 54.4      \\
+ GRPO                   & 50.0        & 10.7     & 46.7        & 11.9     & 91.6       & 55.6     & 65.4        & 55.3      \\
\rowcolor[HTML]{EAE8FD} + GRPO w/ Entropy Adv.           & 53.3        & 11.8     & 56.7        & 12.6     & 91.6       & 57.8     & 74.0        & 58.5      \\
{$\Delta$}                    & \textcolor[HTML]{006400}{+3.3}         & \textcolor[HTML]{006400}{+1.1}      & \textcolor[HTML]{006400}{+10.0}        & \textcolor[HTML]{006400}{+0.7}      & \textcolor{gray}{+0.0}        & \textcolor[HTML]{006400}{+2.2}      & \textcolor[HTML]{006400}{+8.6}         & \textcolor[HTML]{006400}{+3.2}       \\ \cmidrule(lr){1-9}
+ PPO                    & 43.3        & 7.9      & 46.7        & 14.2     & 85.5       & 51.8     & 68.4        & 57.9      \\
\rowcolor[HTML]{EAE8FD} + PPO w/ Entropy Adv.            & 56.7        & 11.7     & 50.0        & 16.8     & 88.0       & 56.1     & 75.2        & 60.9      \\ 
{$\Delta$}             & \textcolor[HTML]{006400}{+13.4}        & \textcolor[HTML]{006400}{+3.8}      & \textcolor[HTML]{006400}{+3.3}         & \textcolor[HTML]{006400}{+2.6}      & \textcolor[HTML]{006400}{+2.5}        & \textcolor[HTML]{006400}{+4.3}      & \textcolor[HTML]{006400}{+6.8}         & \textcolor[HTML]{006400}{+3.0}       \\ \midrule
\hspace{-4pt}\textit{{Qwen2.5-Math}}          & 50.7         & 4.4     & 70.0        & 10.7     & 90        & 34.4     & 88.6        & 47.5      \\
Qwen2.5-Math-Ins\dag &        -       &      -      &         -      & 13.3       &     -         & 50.6       &     -          & 79.8        \\
Eurus-2-PRIME         &        -       &      -      &         -      & 26.7       &     -         & 57.8       &      -         & 79.2        \\
Oat-Zero\dag             &        -       &       -     &       -        & 30.0         &      -        & 55.4       &       -        & 80.6        \\
GPG\dag                   &       -        &       -     &         -      & 33.3       &       -       & 65.0         &       -        & 80.0          \\ \cmidrule(lr){1-9}
+ GRPO                   & 57.4          & 16.3       & 83.3        & 30.9     & 92.8       & 66.9     & 94.6        & 83.0      \\
\rowcolor[HTML]{EAE8FD} + GRPO w/ Entropy Adv.           & 63.6          & 17.6       & 80.0        & 33.7     & 95.2       & 69.8     & 94.8        & 83.1      \\
{$\Delta$}                  & \textcolor[HTML]{006400}{+6.2}           & \textcolor[HTML]{006400}{+1.3}        & \textcolor[HTML]{8B0000}{-3.3}        & \textcolor[HTML]{006400}{+2.8}      & \textcolor[HTML]{006400}{+2.4}        & \textcolor[HTML]{006400}{+2.9}      & \textcolor[HTML]{006400}{+0.2}         & \textcolor[HTML]{006400}{+ 0.1}  \\ \bottomrule    
\end{tabular}%
}
\caption{\textbf{Pass@K and Pass@1 Performance}. \dag: results from~\citet{gpg}. ``+ GRPO'' and ``+ PPO'' indicate RL training from the base models, while ``w/ Entropy Adv.'' denotes applying our entropy-based advantage to the corresponding RL algorithms. $\Delta$ denotes the performance difference between without and with applying our method.}
\label{tab:main_results}
\end{table}

\section{Analysis}
\begin{figure}[!tb]
    \centering
\includegraphics[width=\linewidth]{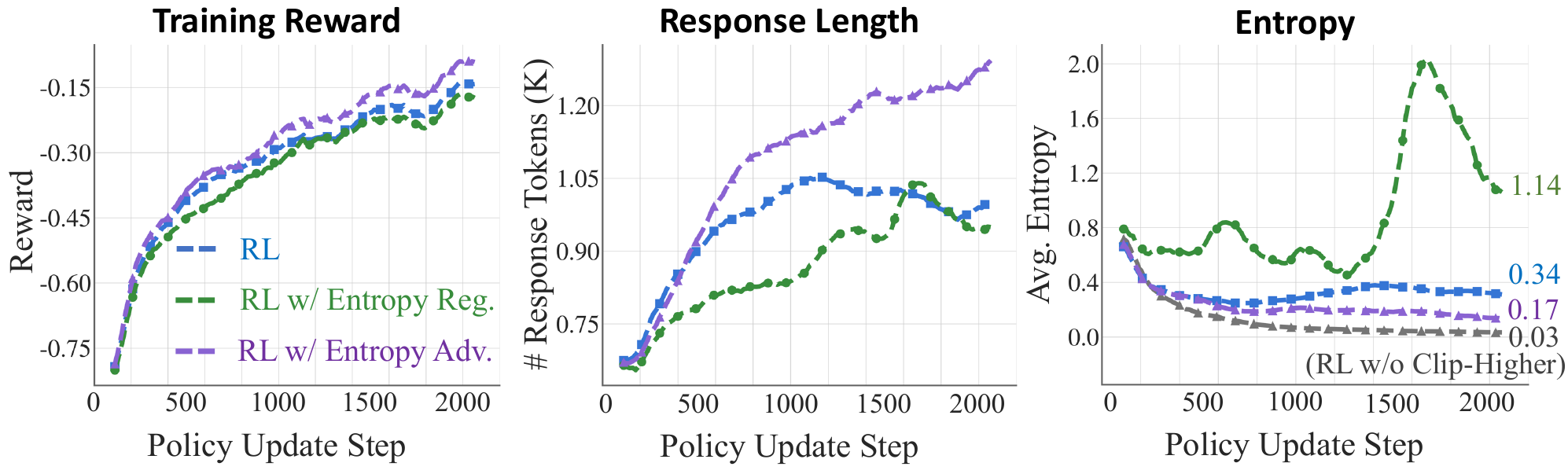}
    \vspace{-17pt}
\caption{\textbf{Metrics during RL training.} The RL baseline is GRPO; ``RL w/ Entropy Reg.'' applies entropy regularization; ``RL w/ Entropy Adv.'' applies entropy-based advantage shaping; ``RL w/o Clip-Higher'' removes the clip-higher technique from the RL baseline (i.e., \( \epsilon_{\mathrm{high}} = \epsilon_{\mathrm{low}} = 0.2 \)). ``Entropy Adv. Ratio'' denotes the ratio of entropy-based advantage to the original advantage.}
\label{fig:train_metrics}
\end{figure}

\begin{figure}[!tb]
    \centering
    \includegraphics[width=\linewidth]{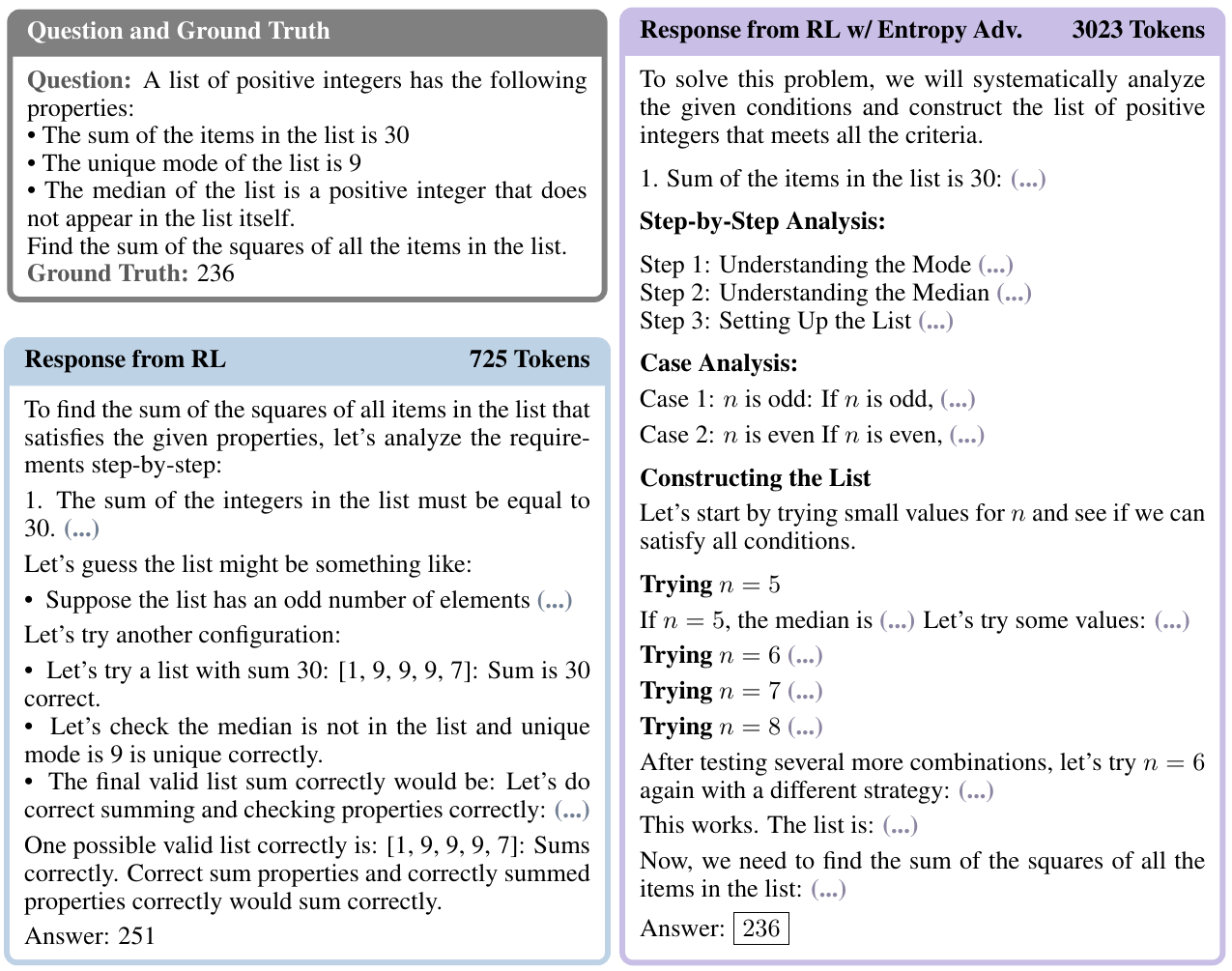}
    \vspace{-15pt}
    \caption{\textbf{Response Comparison} between RL-trained LLMs with and without entropy-based advantage shaping. Certain portions are omitted and the completion version is in Appendix~\ref{app:case study}.}
    \label{fig:case}
    \vspace{-10pt}
\end{figure}

We conduct a detailed analysis to understand the impact of our method on RL training for LLMs. Specifically, we track key metrics—including reward, response length, entropy, and our entropy-based advantage—throughout training (Figure~\ref{fig:train_metrics} and~\ref{fig:bonus_ratio}), as well as reasoning dynamics during testing (Figure~\ref{fig:after_analysis} and~\ref{fig:case}). Furthermore, we provide a comprehensive comparison between our method and traditional entropy regularization.

\subsection{RL Training Process}
\begin{figure}[!tb]
    \centering
    \includegraphics[width=\linewidth]{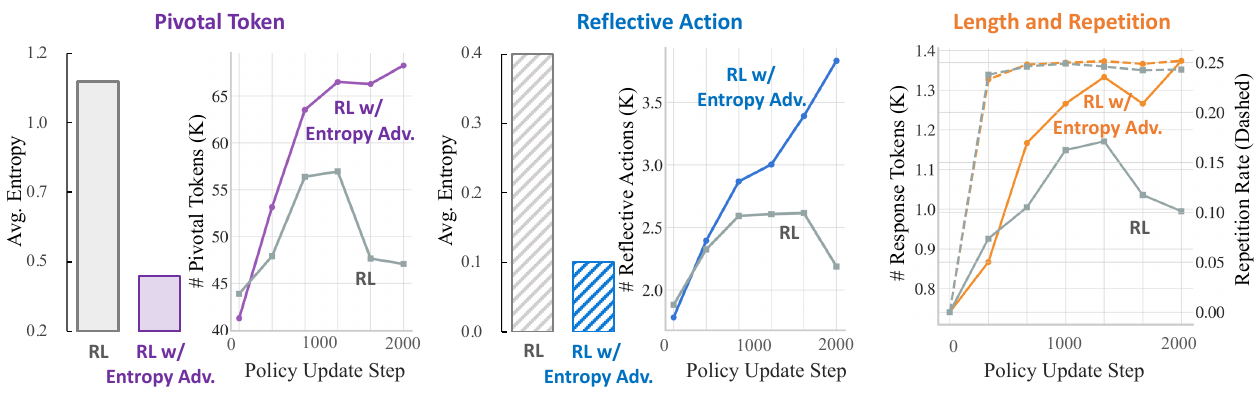}
    \vspace{-20pt}
    \caption{\textbf{Comparison of Pivotal Tokens (left), Reflection Actions (middle), and Response Length and Repetition (right)} on the testing task between the baseline RL algorithm—GRPO (grey)—and RL with entropy advantage (purple/blue/orange), evaluated across training process.}
    \label{fig:after_analysis}
\end{figure}

\paragraph{Training Reward} As shown on the left of Figure~\ref{fig:after_analysis}, we observe steady upward trends across all three methods. Notably, RL with Entropy-based Advantage yields slightly higher rewards in the later stages of training, indicating a stronger and more sustained improvement over time.

\paragraph{Response Length} Figure~\ref{fig:after_analysis} (middle) indicates that while the response length of RL baseline shows a steady increase before 1000 steps, it slightly declines thereafter. In contrast, augmenting the RL baseline with our entropy-based advantage sustains the upward trend in response length beyond 1000 steps, surpassing both the RL baseline and RL with Entropy Regularizer. This may reflect stronger reasoning capabilities, as the LLM tends to spend more time (i.e., generate more tokens) to explore and reach the correct answer~\citep{deepseekr1}.

\paragraph{Overall Entropy} Both the RL baseline and our method exhibit a decreasing trend throughout training, reflecting increasing model confidence. However, neither shows signs of entropy collapse. This is likely due to the use of the ``clip-higher'' technique in the RL baseline, which prevents the gradients of low-probability tokens from being clipped. Specifically, at step 2000, the entropy of the RL baseline is 0.34, and that of our method is 0.17. As a reference, in an ablation where ``clip-higher'' is removed, entropy drops to 0.03—a level typically considered as entropy collapse~\citep{dapo,entropymechanism}. 

\begin{table}[!htb]
\centering
\resizebox{\textwidth}{!}{%
\begin{tabular}{lcccccccc}
\toprule
                         & \multicolumn{2}{c}{AIME25} & \multicolumn{2}{c}{AIME24} & \multicolumn{2}{c}{AMC23} & \multicolumn{2}{c}{MATH500} \\ \cmidrule(lr){2-3} \cmidrule(lr){4-5} \cmidrule(lr){6-7} \cmidrule(lr){8-9}
                         & \textit{Pass@256}      & \textit{Pass@1}     & \textit{Pass@256}      & \textit{Pass@1}     & \textit{Pass@128}     & \textit{Pass@1}     & \textit{Pass@16}       & \textit{Pass@1}      \\ \cmidrule(lr){1-9}

RL w/ Entropy Reg.                   & 50.0 & 9.3 & 50.0 & 16.0 & 90.4 & 54.3 & 70.4 & 57.4     \\
\rowcolor[HTML]{EAE8FD} RL w/ Entropy Adv.           & 53.3        & 11.8     & 56.7        & 12.6     & 91.6       & 57.8     & 74.0        & 58.5  
\\ \bottomrule    
\end{tabular}%
}
\caption{Comparison of model performance trained with RL (i.e., GRPO) using entropy regularization vs. entropy-based advantage shaping}\label{tab:Entropy Reguar}
\vspace{-10pt}
\end{table}

In contrast, although adding an entropy regularizer to the RL training objective increases entropy during training, it shows a sudden spike after step 1500, indicating unstable optimization. The corresponding testing performance comparison between our method and entropy regularization is shown in Table~\ref{tab:Entropy Reguar}, highlighting our method’s superiority in promoting stable training while improving reasoning performance.

\begin{wrapfigure}[14]{r}{0.37\textwidth}
\vspace{-15pt}
\centering
\includegraphics[width=0.37\textwidth]{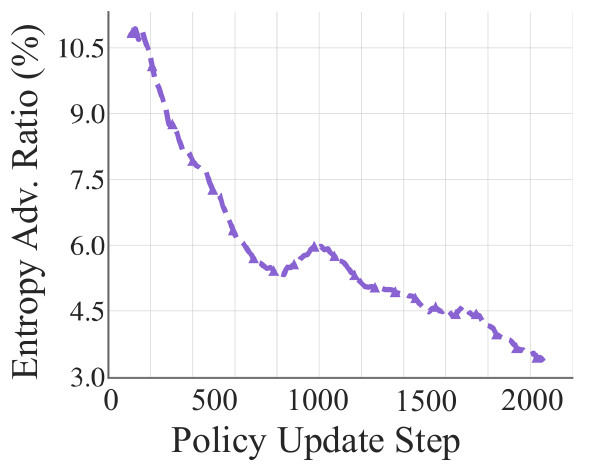}
\vspace{-15pt}
\caption{\textbf{Ratio of Entropy-Based Advantage} to the original advantage (i.e., \( \frac{\psi(\mathcal{H}_t)}{|A_t|} \)).}
\label{fig:bonus_ratio}
\end{wrapfigure}
Our method does not aim to increase token entropy uniformly. Instead, it promotes exploratory reasoning behaviors by amplifying high-entropy actions and gradually builds confidence (reflected by lower entropy) at these points. We further analyze this in the next subsection on exploratory reasoning dynamics.

\paragraph{Entropy-Based Advantage}
Figure~\ref{fig:bonus_ratio} shows the ratio of the entropy-based advantage to the original advantage.
As training progresses and the model gains confidence, the entropy-based advantage decreases. This supports our hypothesis that the intrinsic tension between model confidence and entropy naturally encourages exploration in uncertain regions, while gradually reducing the entropy-based advantage once sufficient confidence is achieved---thereby preventing over-encouragement.

\subsection{Exploratory Reasoning Dynamics}
We further analyze the reasoning dynamics of the RL-trained models on the testing benchmarks to validate whether encouraging high-entropy actions during training can effectively enhance the model’s exploratory reasoning capabilities.

\paragraph{Pivotal Tokens and Reflective Actions}
As shown in Figure~\ref{fig:after_analysis}, applying our entropy-based advantage successfully reinforces the model's ability to generate pivotal tokens and reflective actions. These regions exhibit much lower entropy, indicating increased model confidence when producing such actions. Consequently, we observe significantly higher counts of pivotal tokens and reflective actions compared to the baseline.

\paragraph{Response Length and Repetition Rate}
On the right side of Figure~\ref{fig:after_analysis}, we also observe a substantial increase in response length across testing benchmarks. Additionally, we record the n-gram-based repetition rate of generated responses and find that our method yields much longer responses while maintaining a repetition rate comparable to that of the RL baseline, demonstrating its ability to scale effectively at test time without increasing redundancy.

\paragraph{Case Study}
Figure~\ref{fig:case} presents example responses from the RL-trained models. Compared to the baseline, our method produces more accurate and mathematically rigorous solutions. The model explicitly lists problem constraints, performs systematic case analysis (e.g., odd vs. even list lengths), and dynamically adjusts its approach when initial attempts fail. For instance, it iterates through candidate values (e.g., \(n = 5, 6, \ldots\)) while ensuring constraints are satisfied at each step. This structured and persistent reasoning process leads to valid final answers, whereas the baseline often overlooks key conditions and produces incorrect solutions.

\section{Related Work}
\paragraph{Exploration in Reinforcement Learning} 
Exploration has long been a central theme in RL~\citep{cae}, addressed through theoretical frameworks~\citep{OPPO,PC-PG,LSVI-PHE}, as well as empirical heuristics~\citep{RND,ICM,RIDE,E3B}. Motivated by the use of entropy to guide exploration~\citep{SAC, PPO, InverseRL}, we investigate its role in LLM reasoning by treating entropy as an advantage-shaping signal to reinforce exploratory reasoning behaviors. A concurrent work~\citep{imentor} also studies exploration-driven reasoning but adopts a different approach by designing custom metrics rather than using entropy. Other concurrent studies incorporate an entropy regularizer~\citep{skyworko1,trainoneexample} to the training objective, while our method focuses on the advantage function, providing an orthogonal perspective. 

\paragraph{Training Signals in Reinforcement Fine-Tuning}
Reinforcement fine-tuning of LLMs can leverage supervised and/or unsupervised training signals~\citep{spuriousrewards}. Supervised methods, such as RLHF~\citep{rlhf} and RLVR, rely on reward signals derived from human feedback or verifiable correctness, and have proven effective in aligning model behavior and solving deterministic tasks. In contrast, unsupervised approaches reduce dependence on human annotations by leveraging consistency-based signals~\citep{scpo,ttrl} or entropy minimization~\citep{empo,unreasonableeffectiveness}. \citet{seedgrpo} incorporates uncertainty-aware weighting into RL using semantic entropy. Our work focuses on unsupervised signals with a specific emphasis on exploration, employing entropy to shape the advantage and encourage exploratory reasoning.

\section{Conclusion}
This work investigates reasoning with exploration to encourage longer and deeper reasoning chains in LLMs, through the lens of entropy. We begin by analyzing the relationship between entropy and exploratory reasoning, revealing that pivotal tokens, reflective actions, and rare behaviors consistently align with regions of higher entropy. Motivated by these findings, we introduce a minimal modification to standard RL algorithms by augmenting the advantage function with a clipped, gradient-detached entropy term. This design fosters exploration by promoting longer and deeper reasoning, while preserving the original policy optimization direction. We validate our method on mainstream RLVR algorithms, GRPO and PPO, and observe substantial improvements in Pass@$K$ across diverse benchmarks—highlighting a promising direction for exploration-aware LLM training.

\section*{Acknowledgments}
The first author would like to thank Hongzhao Xie, Yexin Li, and Yuxian Gu for helpful discussions.

\bibliography{ref}
\bibliographystyle{iclr2025_conference}

\clearpage

\appendix
\section{Extended Preliminary Analysis} \label{app:Extended Preliminary Analysis}
\begin{figure}[!htb]
    \centering
    \includegraphics[width=0.6\textwidth]{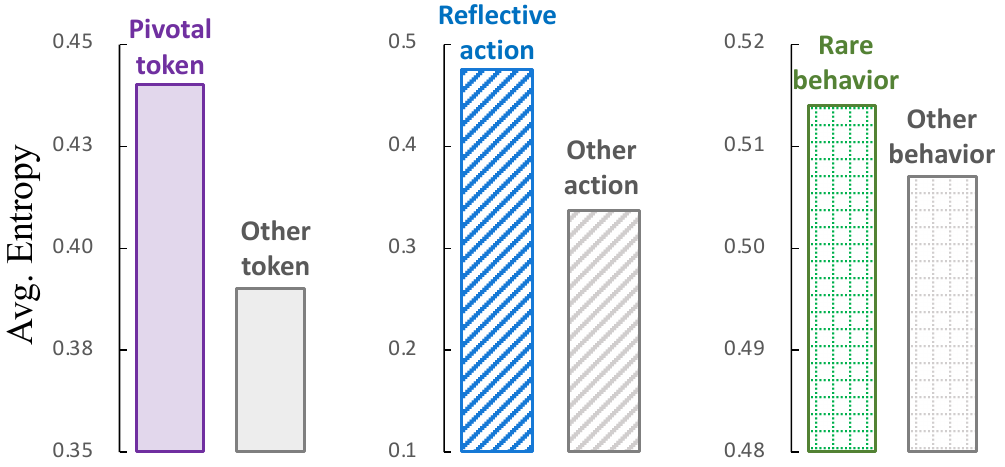}
    \caption{\textbf{Entropy Comparison between Exploratory Reasoning and Others} using DeepSeek-R1-Distill-Qwen-1.5B on code domain.}
    \label{fig:analysis_code}
\end{figure}

Section~\ref{sec:Preliminary Analysis} analyzes Qwen2.5-Base on mathematical domain. We further validate our findings using DeepSeek-R1-Distill-Qwen-1.5B on code domain; the evaluation benchmark is LiveCodeBench~\citep{livecodebench}, and the RL training configuration follows DeepCoder~\citep{deepcoder}. DeepSeek-R1-Distill-Qwen-1.5B is fine-tuned from Qwen2.5-Math-1.5B on large-scale, high-quality long-chain reasoning data distilled from DeepSeek-R1-671B, and already exhibits strong long-chain reasoning capabilities. Figure~\ref{fig:analysis_code} shows consistent results: pivotal tokens, reflective actions, and rare behaviors exhibit higher entropy. This demonstrates that our entropy-related observations hold across domains, even with LLMs that already demonstrate strong reasoning performance.

\section{Experiment Settings}\label{app:Experiment Settings}

\paragraph{Training Data and Reward Function} For Qwen2.5-Base-7B, we use DAPO's original prompt template and verifier; for Qwen2.5-Math-Base-7B, we append the suffix \texttt{\textbackslash nPlease reason step by step, and put your final answer within \textbackslash boxed\{\}.}to each question and employ Math-Verify~\citep{mathverify} as the verifier.

\paragraph{RL Training Configuration} For both GRPO and PPO, we use the hyperparameters in Table~\ref{train_hyperparameters}, without using entropy or KL losses. For PPO only, the critic is trained with a learning rate of $ 2\text{e}^{-6} $, with a 10-step pre-training applied before initiating policy updates. We use GAE with $ \lambda = 1.0 $ and $ \gamma = 1.0 $.

\begin{table}[!htb]
\centering
\begin{tabular}{@{}ll@{}}
\toprule
\textbf{Hyperparameter} & \textbf{Value} \\
\midrule
Optimizer & AdamW \\
Policy learning rate & $1\text{e}^{-6}$ \\
Training batch size & 512 prompts \\
Samples per prompt & 8 \\
Mini-batch size & 32 prompts \\
Policy updates per rollout & 16 \\
Max response length & 8K tokens \\
Rollout temperature & 1.0 \\
Clip range $\epsilon_{\text{low}}$, $\epsilon_{\text{high}}$ & $0.2$, $0.28$ \\
\bottomrule
\end{tabular}
\vspace{-5pt}
\caption{RL Hyperparameters}\label{train_hyperparameters}
\end{table}

\section{Case Study}
\label{app:case study}
\begin{tcolorbox}[
    title=Question and Ground Truth,
    colback=white,
    colframe=gray,
    coltitle=white,
    fonttitle=\bfseries,
    arc=1mm,
    boxrule=0.6mm,
    left=1mm,   
    right=1mm,  
    top=1mm,     
    bottom=1mm, 
]
\textcolor{gray!70!black}{\textbf{Question:}} A list of positive integers has the following properties: (1) The sum of the items in the list is 30; (2) The unique mode of the list is 9; (3) The median of the list is a positive integer that does not appear in the list itself. Find the sum of the squares of all the items in the list.

\textcolor{gray!70!black}{\textbf{Ground Truth:}} 236
\end{tcolorbox}

\begin{tcolorbox}[
    title=Response from RL Baseline~~~~~~~~~~~~~~~~~~~~~~~~~~~~~~~~~~~~~~~~~~~~~~~~~~~~~~~~~~~~~~~Total Length: 725 Tokens,
    colback=white,
    colframe=lightblue,
    coltitle=black,
    fonttitle=\bfseries,
    arc=1mm,
    boxrule=0.6mm,
    left=1mm,    
    right=1mm,   
    top=1mm,     
    bottom=1mm,  
]
To find the sum of the squares of all items in the list that satisfies the given properties, let's analyze the requirements step-by-step:
\\[3pt]
1. The sum of the integers in the list must be equal to 30.

2. The unique mode of the list is 9.

3. The median of the integer list is a positive integer that does not appear in the list.
\\[6pt]
Let's denote the list as $ a_1, a_2, \ldots, a_n $ where $ a_1 \leq a_2 \leq \cdots \leq a_n $.
\\[6pt]
The mode of the list being unique and equal to 9 implies that 9 must appear the most frequently in the list; no other integer in the list appears as frequently as 9.
\\[6pt]
The median of the list is a positive integer not found in the list. Depending on whether the list has an odd or even number of elements, the median will be the middle element in the sorted list or the average of two middle elements, respectively.
\\[6pt]
Now let's consider the possible list structures based on these properties. Since 9 must be the unique mode, 9 must appear more frequently than any other integer in the list. Let's consider the possible sizes of the list and the appearances of 9.
\\[6pt]
Let's guess the list might be something like:
\\[3pt]
•~~~Suppose the list has an odd number of elements, let $ n = 2k + 1 $. The median would be the element in the middle position.

•~~~Suppose the median is $ m $, and $ m \neq a_1, a_2, \ldots, a_n $.
\\[6pt]
Let's consider the list with these properties:
\\[3pt]
•~~~The list might include several 9's. Let's start by considering the minimal configurations:
\\
•~~~Suppose the list has exactly three 9's because then 9 appears with the highest frequency and other numbers must sum correctly. Thus the list will look like: 9, 9, 9, \ldots along with other integers summing up correctly.
\\[6pt]
We need to find a configuration where the sum is 30, and the median does not appear, unique mode is 9, and the median is integers in the list.
\\[6pt]
Let's try constructing such a list:
\\[3pt]
•~~~Let's say the list has a number of 9's (let's assume $ 3 \times 9 = 27 $), then we need the rest to sum to $ 3 \leq 3 $.

•~~~Let's try constructing a specific list configuration manually. Suppose the list is; [1, 9, 9, 9, 7]:

~~~~~Sum is $ 1 + 9 + 9 + 9 + 7 = 35 $ which is not correct.
\\[6pt]
Let's try another configuration:
\\[3pt]
•~~~Let's try a list with sum 30: [1, 9, 9, 9, 7]: Sum is 30 correct.

•~~~Let's check the median is not in the list and unique mode is 9 is unique correctly.

•~~~The final valid list sum correctly would be: Let's do correct summing and checking properties correctly: 

Sum correctly would be 30:
\\
~~~~~One possible valid list correctly is: [1, 9, 9, 9, 7]: Sums correctly. Correct sum properties and correctly summed properties correctly would sum correctly.

Answer: 251
\end{tcolorbox}

\begin{tcolorbox}[
    title=Response from RL w/ Entropy Advantage (Part 1)~~~~~~~~~~~~~~~~~~~~~~~~Total Length: 3023 Tokens,
    colback=white,
    colframe=lavender,
    coltitle=black,
    fonttitle=\bfseries,
    arc=1mm,
    boxrule=0.6mm,
    left=1mm,    
    right=1mm,   
    top=1mm,     
    bottom=1mm,  
]
To solve this problem, we will systematically analyze the given conditions and construct the list of positive integers that meets all the criteria.
\\[6pt]
\textbf{1. Sum of the items in the list is 30:}
$$
\text{Sum} = 30
$$
\textbf{2. The unique mode of the list is 9:}
$$
\text{Mode} = 9 \quad \text{and} \quad 9 \text{~appears more frequently than any other number}
$$
\textbf{3. The median of the list is a positive integer that does not appear in the list itself:}
$$
\text{Median is a positive integer} \quad \text{and} \quad \text{Median} \notin \text{list}
$$
Let's start by denoting the list of positive integers as $ a_1, a_2, \ldots, a_n $ where $ n $ is the number of elements in the list, and the list is sorted in ascending order ($ a_1 \leq a_2 \leq \cdots \leq a_n $).
\\
\\
{\textbf{Step-by-Step Analysis:}}
\\[6pt]
\textbf{Step 1: Understanding the Mode}
\\[3pt]
Since the mode is 9 and it is unique, 9 must appear more frequently than any other number in the list. Let's say 9 appears $ k $ times. Therefore, $ k $ is the highest frequency of any number in the list.
\\[3pt]
\textbf{Step 2: Understanding the Median}
\\[3pt]
The median is the middle value of the list when it is sorted. Since the median is a positive integer that does not appear in the list, it must be the average of the two middle numbers if $ n $ is even, or it must be the middle number if $ n $ is odd.
\\[3pt]
\textbf{Step 3: Setting Up the List}
\\[3pt]
Let's start by assuming the smallest possible values for the other numbers while maintaining the mode as 9.
\\
\\
\textbf{Case Analysis:}
\\[6pt]
\textbf{Case 1: $ n $ is odd}
\\[3pt]
If $ n $ is odd, the median is the middle number $ a_{\frac{n+1}{2}} $. Since the median is a positive integer not in the list, $ a_{\frac{n+1}{2}} \neq 9 $.
\\[3pt]
\textbf{Case 2: $ n $ is even}
\\[3pt]
If $ n $ is even, the median is the average of the two middle numbers $ \frac{a_{\frac{n}{2}} + a_{\frac{n}{2}+1}}{2} $. Since the median is a positive integer not in the list, $ \frac{a_{\frac{n}{2}} + a_{\frac{n}{2}+1}}{2} \neq 9 $.
\\
\\
\textbf{Constructing the List}
\\[3pt]
Let's start by trying small values for $ n $ and see if we can satisfy all conditions.
\\[6pt]
\textbf{Trying $ n = 5 $}
\\[3pt]
If $ n = 5 $, the median is $ a_3 $. Since $ a_3 $ is a positive integer not in the list, $ a_3 \neq 9 $. The list looks like: $ a_1, a_2, 9, a_4, a_5 $
\\[3pt]
The sum is:
$$
a_1 + a_2 + 9 + a_4 + a_5 = 30
$$
Since 9 is the mode, $ a_1, a_2, a_4, $ and $ a_5 $ cannot all be 9 (as 9 would not be the unique mode). Let's try some values:
\\[3pt]
•~~~If $ a_1 = a_2 = a_4 = a_5 = 1 $ (smallest possible values):
$$
1 + 1 + 9 + 1 + 1 = 13 \quad (\text{not equal to } 30)
$$
•~~~If $ a_1 = a_2 = a_4 = 1 $ and $ a_5 = 2 $:
$$
1 + 1 + 9 + 1 + 2 = 14 \quad (\text{not equal to } 30)
$$
\end{tcolorbox}

\begin{tcolorbox}[
    title=Response from RL w/ Entropy Advantage (Part 2)~~~~~~~~~~~~~~~~~~~~~~~~Total Length: 3023 Tokens,
    colback=white,
    colframe=lavender,
    coltitle=black,
    fonttitle=\bfseries,
    arc=1mm,
    boxrule=0.6mm,
    left=1mm,    
    right=1mm,   
    top=1mm,     
    bottom=1mm,  
]
•~~~If $ a_1 = a_2 = 1 $ and $ a_4 = a_5 = 2 $:
$$
1 + 1 + 9 + 2 + 2 = 15 \quad (\text{not equal to } 30)
$$
•~~~If $ a_1 = 1 $ and $ a_2 = a_4 = a_5 = 2 $:
$$
1 + 2 + 9 + 2 + 2 = 16 \quad (\text{not equal to } 30)
$$

We need to find a better combination. Let's try increasing the values while maintaining the mode and unique mode properties.
\\
\\
\textbf{Trying $ n = 6 $}
\\[3pt]
If $ n = 6 $, the median is the average of $ a_3 $ and $ a_4 $. Since the median is a positive integer not in the list, $ \frac{a_3 + a_4}{2} \neq 9 $. The list looks like: $ a_1, a_2, a_3, 9, 9, a_6 $
\\[3pt]
The sum is:
$$
a_1 + a_2 + a_3 + 9 + 9 + a_6 = 30
$$

Since 9 is the mode, $ a_1, a_2, a_3, $ and $ a_6 $ cannot all be different. Let's try some values:

•~~~If $ a_1 = a_2 = 1 $ and $ a_3 = a_6 = 2 $:
$$
1 + 1 + 2 + 9 + 9 + 2 = 24 \quad (\text{not equal to } 30)
$$
•~~~If $ a_1 = 1 $ and $ a_2 = a_3 = 2 $ and $ a_6 = 3 $:
$$
1 + 2 + 2 + 9 + 9 + 3 = 26 \quad (\text{not equal to } 30)
$$
•~~~If $ a_1 = 1 $ and $ a_2 = 2 $ and $ a_3 = a_6 = 3 $:
$$
1 + 2 + 3 + 9 + 9 + 3 = 27 \quad (\text{not equal to } 30)
$$
•~~~If $ a_1 = 2 $ and $ a_2 = 2 $ and $ a_3 = a_6 = 3 $:
$$
2 + 2 + 3 + 9 + 9 + 3 = 28 \quad (\text{not equal to } 30)
$$
•~~~If $ a_1 = 1 $ and $ a_2 = 3 $ and $ a_3 = 5 $ and $ a_6 = 6 $:
$$
1 + 3 + 5 + 9 + 9 + 6 = 33 \quad (\text{not equal to } 30)
$$

After testing several combinations, let's try a different approach by starting with a known mode and ensuring the sum is met.
\\
\\
\textbf{Trying $ n = 7 $}
\\[3pt]
If $ n = 7 $, the median is $ a_4 $. Since the median is a positive integer not in the list, $ a_4 \neq 9 $. The list looks like: $ a_1, a_2, a_3, 9, 9, 9, a_7 $
\\[3pt]
The sum is:
$$
a_1 + a_2 + a_3 + 9 + 9 + 9 + a_7 = 30
$$

Since 9 is the mode, we need to balance the remaining values. Let's try:
\\
•~~~$ a_1 = a_2 = a_3 = 1 $:
$$
1 + 1 + 1 + 9 + 9 + 9 + a_7 = 30
$$
$$
30 + a_7 = 30
$$
$$
a_7 = 0 \quad (\text{not valid since } a_7 \text{ must be positive})
$$
•~~~$ a_1 = a_2 = 1 $ and $ a_3 = 2 $:
$$
1 + 1 + 2 + 9 + 9 + 9 + a_7 = 30
$$
$$
31 + a_7 = 30
$$
$$
a_7 = -1 \quad (\text{not valid since } a_7 \text{ must be positive})
$$
After testing several more combinations, let's try a different approach by starting with a known mode and ensuring the sum is met while satisfying all conditions.
\end{tcolorbox}

\begin{tcolorbox}[
    title=Response from RL w/ Entropy Advantage (Part 3)~~~~~~~~~~~~~~~~~~~~~~~~Total Length: 3023 Tokens,
    colback=white,
    colframe=lavender,
    coltitle=black,
    fonttitle=\bfseries,
    arc=1mm,
    boxrule=0.6mm,
    left=1mm,    
    right=1mm,   
    top=1mm,     
    bottom=1mm,  
]
\textbf{Trying $ n = 8 $}
\\[3pt]
If $ n = 8 $, the median is the average of $ a_4 $ and $ a_5 $. Since the median is a positive integer not in the list, $ \frac{a_4 + a_5}{2} \neq 9 $. The list looks like: $ a_1, a_2, a_3, 9, 9, 9, a_6, a_7 $
\\[3pt]
The sum is:
$$
a_1 + a_2 + a_3 + 9 + 9 + 9 + a_6 + a_7 = 30
$$

Since 9 is the mode, we need to balance the remaining values. Let's try:

•~~~$ a_1 = a_2 = a_3 = 1 $:
$$
1 + 1 + 1 + 9 + 9 + 9 + a_6 + a_7 = 30
$$
$$
3 + 27 + a_6 + a_7 = 30
$$
$$
30 + a_6 + a_7 = 30
$$
$$
a_6 + a_7 = 0 \quad (\text{not valid since } a_6 \text{ and } a_7 \text{ must be positive})
$$
After testing several more combinations, let's try $ n = 6 $ again with a different strategy:
\\[3pt]
•~~~Let's try $ a_1 = a_2 = 1 $, $ a_3 = 2 $, $ a_4 = 9 $, and $ a_5 $ and $ a_6 $ need to be chosen such that the sum is 30 and 9 is the unique mode. The list looks like: $ 1, 1, 2, 9, a_5, a_6 $

The sum is:
$$
1 + 1 + 2 + 9 + a_5 + a_6 = 30
$$
$$
13 + a_5 + a_6 = 30
$$
$$
a_5 + a_6 = 17
$$

Since 9 is the unique mode, $ a_5 $ and $ a_6 $ cannot be 9. Let's try other values:
\\[3pt]
•~~~If $ a_5 = 7 $ and $ a_6 = 10 $:
$$
1 + 1 + 2 + 9 + 7 + 10 = 30
$$
~~~This works. The list is: $ 1, 1, 2, 9, 7, 10 $

~~~•~~~Mode is 9 (unique)

~~~•~~~Median is $ \frac{2 + 7}{2} = 4.5 $ (not in list, integer condition satisfied)
\\[3pt]
Now, we need to find the sum of the squares of all the items in the list:
$$
1^2 + 1^2 + 2^2 + 9^2 + 7^2 + 10^2
$$
$$
1 + 1 + 4 + 81 + 49 + 100
$$
$$
1 + 1 + 4 + 81 + 49 + 100 = 236
$$

Answer: $\boxed{236}$
\end{tcolorbox}

\end{document}